# Discrete-time Robust PD Controlled System with DOB/CDOB Compensation for High Speed Autonomous Vehicle Path Following

**Haoan Wang, Levent Guvenc**

Automated Driving Lab, Ohio State University

## Abstract

In recent years, there has been increasing research on automated driving technology. Autonomous vehicle path following performance is one of significant consideration. This paper presents discrete time design of robust PD controlled system with disturbance observer (DOB) and communication disturbance observer (CDOB) compensation to enhance autonomous vehicle path following performance. Although always implemented on digital devices, DOB and CDOB structure are usually designed in continuous time in the literature and also in our previous work. However, it requires high sampling rate for continuous-time design block diagram to automatically convert to corresponding discrete-time controller using rapid controller prototyping systems. In this paper, direct discrete time design is carried out. Digital PD feedback controller is designed based on the nominal plant using the proposed parameter space approach. Zero order hold method is applied to discretize the nominal plant, DOB and CDOB structure in continuous domain. Discrete time DOB is embedded into the steering to path following error loop for model regulation in the presence of uncertainty in vehicle parameters such as vehicle mass, vehicle speed and road-tire friction coefficient and rejecting external disturbance like crosswind force. On the other hand, time delay from CAN bus based sensor and actuator command interfaces results in degradation of system performance since large negative phase angles are added to the plant frequency response. Discrete time CDOB compensated control system can be used for time delay compensation where the accurate knowledge of delay time value is not necessary. A validated model of our lab's Ford Fusion hybrid automated driving research vehicle is used for the simulation analysis while the vehicle is driving at high speed. Simulation results successfully demonstrate the improvement of autonomous vehicle path following performance with the proposed discrete time DOB and CDOB structure.

## I. Introduction

During the past decades, autonomous vehicle driving technology has been developing rapidly. Researchers are investigating different steering control methods to improve path following performance of autonomous vehicle. In [1], a double loop PD-PID controller is designed for the vehicle steering control. The inner loop is a PID controller which performs to control the position of steering wheel while the outer loop is a PD controller which aims at vehicle's heading control. Similar to double loop PD-PID controller, nested PI and PID controller was proposed in [2]. PI steering controller at inner loop that reduces yaw rate tracking error is used to improve the vehicle steering dynamics. A PID controller is employed at external control loop to reject the lateral deviation from the desired path due to road curvature disturbance. Sliding model control has been widely used due to its benefits of fast and good transient response and robustness with respect to system uncertainties and external disturbances [3]. As another classic control method, model predictive control has the capability to deal with a wide variety of process control constraints systematically and is applied to the socially acceptable collision free path following system in [4]. Parameter space approach based robust PID controller design is presented in [5-6] and it has the advantage of dealing with variable vehicle parameters such as vehicle mass, vehicle velocity and road-tire friction coefficient.

In order to further improve autonomous vehicle path following performance in the existence of uncertain parameters and external disturbance, a disturbance observer (DOB) is added into the control system to achieve insensitivity to modeling error and disturbance rejection. The disturbance observer was firstly proposed by Ohnishi [7] and further developed by Umeno and Hori [8]. Later, DOB has been applied in mechatronic applications in the literature. In [9], robustness of disturbance observer is added to the model of electrohydraulic system considering the case in which the plant has large parametric variation. Two-degrees-of-freedom control architecture known as the model regulator (disturbance observer) is proposed in [10] as a robust steering controller for improving yaw stability in a driver-assist system.

In the autonomous vehicle path following system, CAN bus delay in the steering system is another important issue. Time delay causes large negative phase angles which lead to performance degradation or even instability of the system. The Smith predictor has been firstly introduced and used in many different cases such as [11-12]. It has the advantage of easy implementation. However, time delay model and model accuracy in the knowledge of time delay are required to ensure no degradation of compensation performance. Communication disturbance observer is proposed as another time delay compensation approach. It was firstly applied in the bilateral teleoperation systems [13] and has been extended to robust time delayed control system in [14-15]. Compared with Smith predictor, the accurate knowledge of the time delay value is not necessary in communication disturbance observer and also it can be used for plants with variable time delay.

Disturbance observer and communication disturbance observer are usually designed in continuous time domain and conducted on the digital platform using a very high sampling rate. This requires high speed processors and may not be achievable in many situations.

Therefore, it is worthwhile to investigate direct discrete time design for DOB and CDOB compensated system. [16] discusses application of different discretization methods in discrete implementation of the DOB based control and analyzes three very popular discretization methods: backward difference, bilinear transform, and Al-Alaoui method. It shows that bilinear transform method and Al-Alaoui method provide significantly better performance than backward difference method. A state-space analysis of discrete-time DOB for a class of sampled-data control systems is presented in [17], where discrete-time singular perturbation theory is used to make uncertain sampled-data control system with the discrete-time DOB behaves as the nominal model without disturbance. [18] analysis robust stability condition for discrete time DOB designed by using forward difference discretization method where it is observed that the ratio between time constant of $Q$ filter and sampling time is of significant importance in discrete time DOB and the ratio is suggested to be one for stability. In [19], a discrete-time communication disturbance observer is applied in a network-based gait rehabilitation system for compensating time delay which exist in both sensor-controller and controller actuator channels.

This paper is an extension of our previous work about DOB and CDOB compensated autonomous vehicle path following control system from continuous time domain to discrete time domain. Digital robust PD controller is designed based on the parameter space approach. Uncertainty box illustrating vehicle parameter variations is formed where the vehicle is operating at high speed. DOB structure and CDOB structure are discretized using zero-order-hold method. Simulation results show that discrete time DOB compensated system deals with model regulation and external disturbance rejection and discrete time CDOB compensated system realizes time delay compensation. Both of them present better path following performance than PD feedback controlled system.

The rest of this paper is organized as follows. Section II presents the validated single track vehicle model used for the autonomous vehicle path following. The structures of disturbance observer and communication disturbance observer in discrete time domain are illustrated in Section III and Section IV, respectively. Section V presents parameter space approach based digital robust PD controller design and discretization of DOB and CDOB structures through ZOH method is given in section VI. Section VII shows simulation results of autonomous vehicle path following using discrete time PD with DOB compensation and CDOB compensation, respectively. The paper ends with conclusions in Section VIII.

## II. Vehicle Model

A single track vehicle model presented in Figure 1 is used to model the steering dynamics. The state space model can be described as:

$$\begin{bmatrix} \dot{\beta} \\ \dot{r} \\ \Delta\dot{\psi} \\ \dot{y} \end{bmatrix} = \begin{bmatrix} a_{11} & a_{12} & 0 & 0 \\ a_{21} & a_{22} & 0 & 0 \\ 0 & 1 & 0 & 0 \\ V & l_s & V & 0 \end{bmatrix} \begin{bmatrix} \beta \\ r \\ \Delta\psi \\ y \end{bmatrix} + \begin{bmatrix} b_{11} \\ b_{21} \\ 0 \\ 0 \end{bmatrix} \delta_f + \begin{bmatrix} 0 \\ 0 \\ -\mathrm{V} \\ -l_s\mathrm{V} \end{bmatrix} \rho_{ref} \quad (1)$$

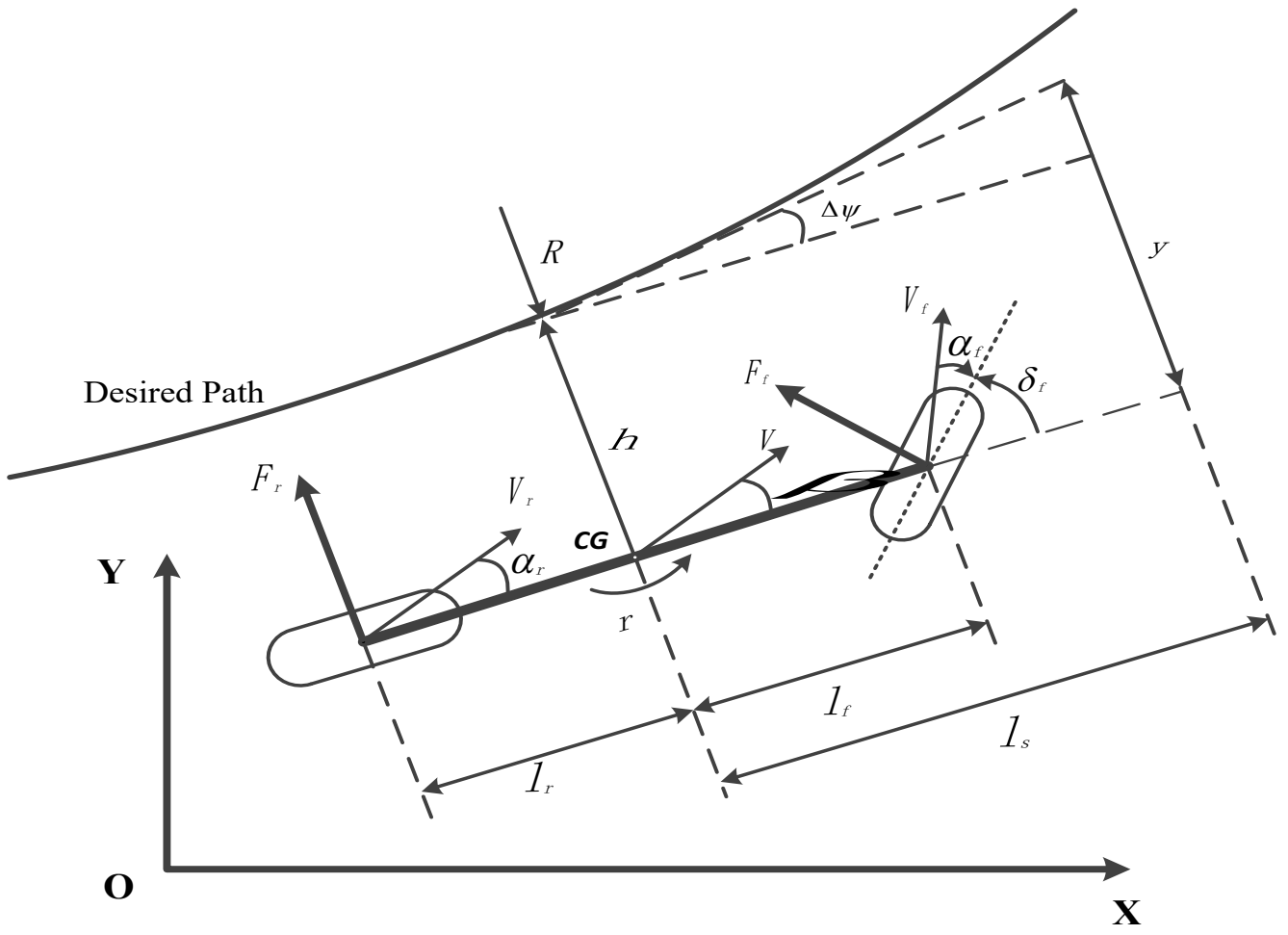

Figure 1. Single track vehicle model diagram

As crosswind force has some influence on dynamics of the vehicle system, the extended vehicle model is given as:

$$\begin{bmatrix} \dot{\beta} \\ \dot{r} \\ \Delta\dot{\psi} \\ \dot{y} \end{bmatrix} = \begin{bmatrix} a_{11} & a_{12} & 0 & 0 \\ a_{21} & a_{22} & 0 & 0 \\ 0 & 1 & 0 & 0 \\ V & l_s & V & 0 \end{bmatrix} \begin{bmatrix} \beta \\ r \\ \Delta\psi \\ y \end{bmatrix} + \begin{bmatrix} b_{11} \\ b_{21} \\ 0 \\ 0 \end{bmatrix} \delta_f + \begin{bmatrix} 0 \\ 0 \\ -\mathrm{V} \\ -l_s\mathrm{V} \end{bmatrix} \rho_{ref} + \begin{bmatrix} \frac{1}{mV} \\ 0 \\ 0 \\ 0 \end{bmatrix} F_{wind} + \begin{bmatrix} 0 \\ \frac{l_{wind}}{J} \\ 0 \\ 0 \end{bmatrix} F_{wind} \quad (2)$$

where

$$a_{11}\text{=-}(c_r\text{+}c_f)/\tilde{m}V,\ a_{12}\text{=-1+}(c_r l_r - c_f l_f)/\tilde{m}V^2 \quad (3)$$

$$a_{21}\text{=}(C_r l_r - c_f l_f)/J,\ a_{22}\text{=}-(c_r l_r^2 + c_f l_f^2)/JV^2$$

$$b_{11}\text{=}c_f/\tilde{m}V,\ b_{21}\text{=}c_f lf/J$$

Writing vehicle steering dynamics in standard form according to (2):

$$\dot{x} = Ax + Bu \quad (4)$$

The transfer function from front wheel steering angle $\delta_f$ to the lateral deviation $y$ in continuous time domain is calculated as equation (5):

$$\frac{y}{\delta_f} = G_{\delta_f} = [0\ 0\ 0\ 1](sI - A)^{-1} \begin{bmatrix} b_{11} \\ b_{21} \\ 0 \\ 0 \end{bmatrix} \quad (5)$$

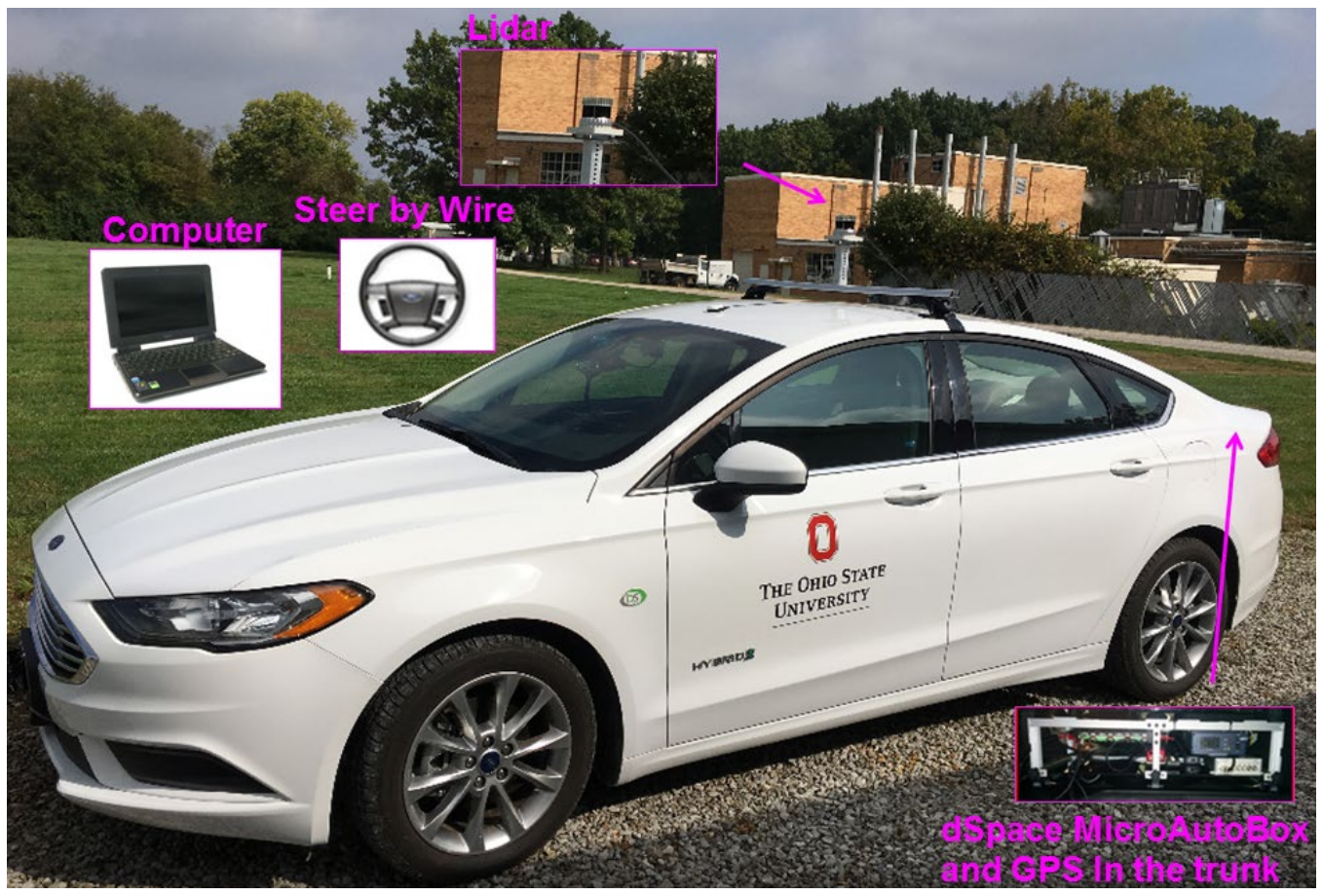

Figure 2. Experiment vehicle

A Ford Fusion hybrid sedan shown in Figure 2 is used as the vehicle under consideration for controller design and simulation. The values of single track model parameters given in Table 1 are measured from experimental vehicle and the vehicle model is validated as seen in reference [20]. Vehicle virtual mass $\tilde{m}(m/\mu)$, vehicle velocity $V$ and road friction coefficient $\mu$ are taken as three uncertain parameters of interest. The nominal values of these three parameters are 2,000 kg, 60 km/hr and 1, respectively. The uncertainty box showing maximum parametric variation in velocity $V$ and virtual mass $\tilde{m}$ of this vehicle is shown in Figure 3. Four vertices labeled by a, b, c, d in the uncertainty box are used to evaluate the performance improvement of the disturbance observer compensated system.

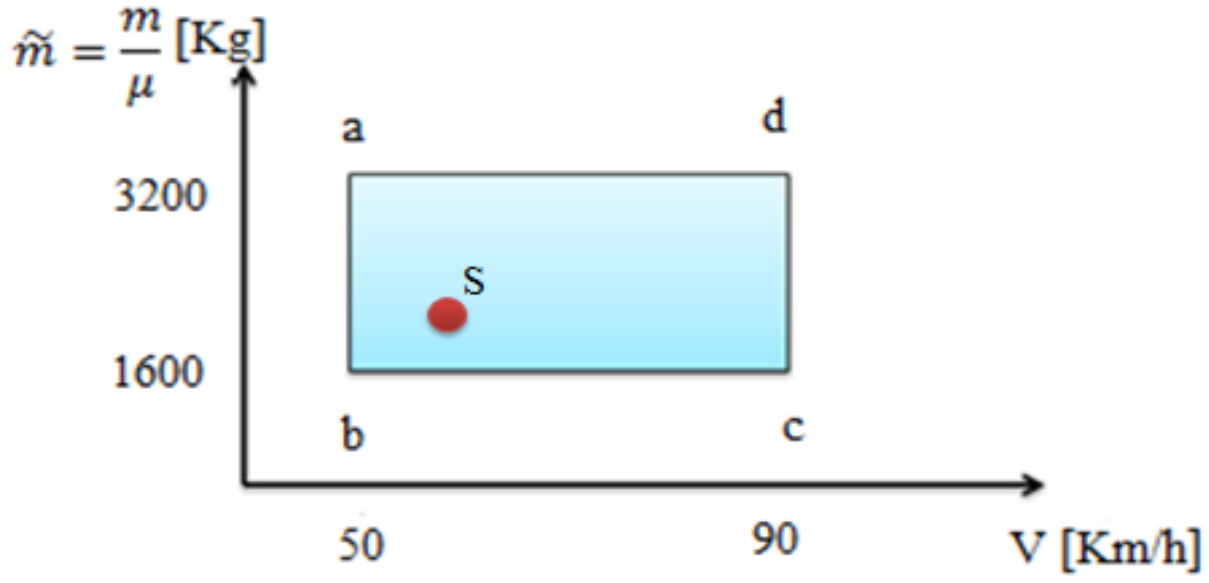

Figure 3. Parametric Uncertainty Box

Table I. Ford Fusion vehicle model parameters

| | |
|---|---|
| $\beta$ | vehicle side slip angle [rad] |
| subscript $f$ | front tires |
| $V$ | vehicle velocity [m/s] |
| $\delta_f$ | front wheel steering angle [rad] |
| $J$ | yaw moment of inertia [3728 kgm$^2$] |
| $C_r$ | rear cornering stiffness [50,000 N/rad] |
| $l_f$ | distance from CG to front axle [1.3008 m] |
| $l_r$ | distance from CG to rear axle [1.5453 m] |
| $\rho_{ref}=1/R$ | curvature of path [1/m] |
| $r$ | vehicle yaw rate [rad/s] |
| subscript $r$ | rear tires |
| $\Delta\psi$ | yaw orientation error with respect to path [rad] |
| $y$ | lateral deviation [m] |
| $C_f$ | front cornering stiffness [195,000 N/rad] |
| $m$ | vehicle mass [2,000 kg] |
| $l_s$ | preview distance [2m] |
| $F_{wind}$ | crosswind force [500N] |
| $l_{wind}$ | horizontal distance from crosswind force acting point to vehicle's $CG$ |

## III. Discrete Time Disturbance Observer

The block diagram of the discrete time closed-loop control system with disturbance observer compensation is depicted in Figure 4. In the block diagram, robust digital PD feedback controller is used as a baseline controller which is designed based on the nominal discrete model of the vehicle. $Q(z)$ is the low pass filter to be selected and its bandwidth determines the bandwidth of model regulation and disturbance rejection. System plant $G(z)$ is formulated by taking both model uncertainty $\Delta_m(z)$ and external disturbance $d$ into account. The vehicle input - output relation becomes:

$$y = G(z)u + d = (G_n(z)(1 + \Delta_m(z)))u + d \qquad (6)$$

where $G_n(z)$ is the desired model of plant and $G(z)$ represents the actual plant. The goal in disturbance observer design is to obtain:

$$y = G_n(z)u_1 \qquad (7)$$

as the input-output relation in the presence of model uncertainty $\Delta_m(z)$ and external disturbance $d$. $u_1$ is regarded as a new steering input which is derived as follows. By considering model uncertainty and external disturbance as an extended disturbance $e$, equation (6) can be rewritten as (8):

$$y = (G_n(z)(1 + \Delta_m(z)))u + d = G_n(z)u + e \qquad (8)$$

Combining equation (7) with equation (8), the new control input $u_1$ is represented as:

$$u_1 = u + \frac{e}{G_n(z)} \quad (9)$$

and

$$u = u_1 - \frac{e}{G_n(z)} = u_1 - \frac{y}{G_n(z)} + u \quad (10)$$

In order to limit the compensation to a low frequency range to avoid stability robustness problem at high frequency, the feedback signals in (10) are multiplied by the low pass filter *Q(z)* and implementation equation becomes:

$$u = u_1 - \frac{Q}{G_n(z)} y + Qu \quad (11)$$

According to discrete time disturbance observer structure in Figure 4, we have the following equation (*z* is omitted in transfer function for simplicity):

$$Y(z) = \frac{CGG_n}{G_n(1-Q)+G(CG_n+Q)} R(z) + \frac{G_n(1-Q)}{G_n(1-Q)+G(CG_n+Q)} D(z) \quad (12)$$

In DOB, $Q$ is chosen as a unity low pass filter, with $Q \rightarrow 1$,

$$\frac{Y(z)}{R(z)} = \frac{CGG_n}{G_n(1-Q)+G(CG_n+Q)} \rightarrow \frac{CG_n}{(1+CG_n)} \quad (13)$$

From (13), we can see that DOB augmented plant behaves like its nominal plant $G_n$, which realizes the model regulation. Also, with $Q \rightarrow 1$,

$$\frac{Y(z)}{D(z)} = \frac{G_n(1-Q)}{G_n(1-Q)+G(CG_n+Q)} \rightarrow 0 \quad (14)$$

which realizes disturbance rejection.

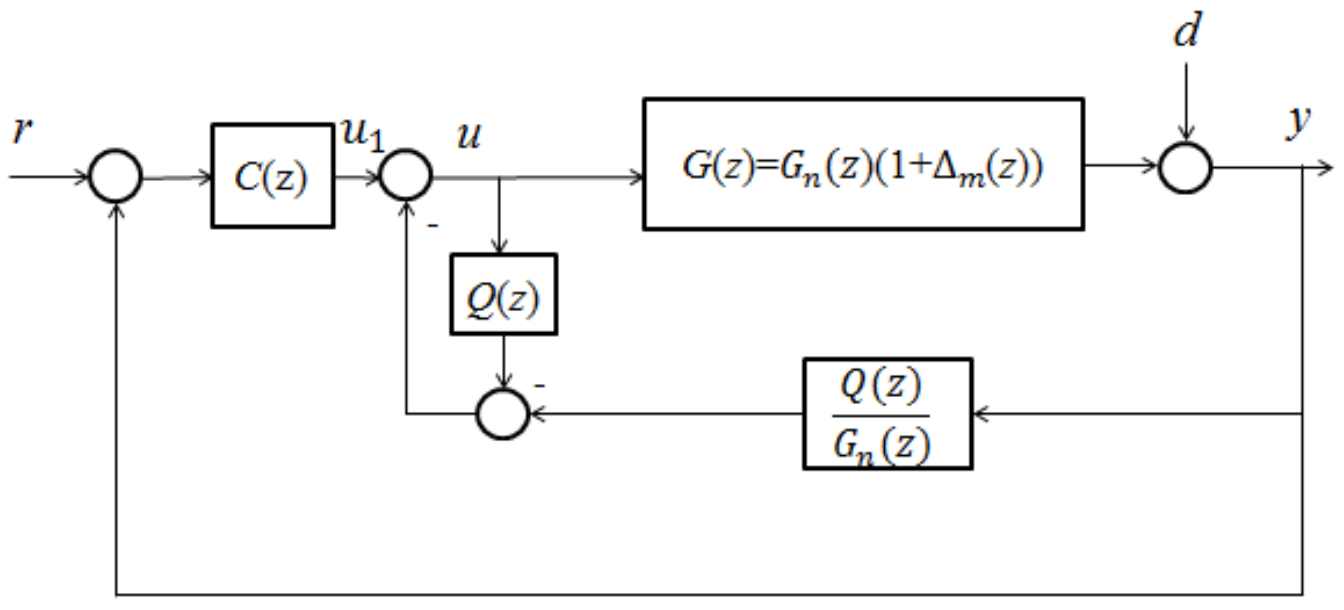

Figure 4. Digital Disturbance Observer Compensated Control System

## IV. Discrete Time Communication Disturbance Observer

In order to deal with path following performance degradation or even destabilization caused by the time delay from CAN bus based sensor and actuator command interfaces, communication disturbance observer is applied to compensate the time delay. Figure 5 illustrates block diagram of time delay estimation for CDOB design in *z* domain. Time delay is considered as a disturbance *d* which is injected on the system and the aim is to obtain disturbance estimation $\hat{d}$. Equation (15) is obtained from Figure 5 and it can be rewritten as (16). Then, the estimated disturbance $\hat{d}$ is derived by multiplying $d$ with *Q*(*z*) to ensure causality as shown in equation (17).

$$y = G_n(z)(u - d) \quad (15)$$

$$d = u - G_n(z)^{-1} y \quad (16)$$

$$\hat{d} = Q(z)\,(u - G_n(z)^{-1} y) \quad (17)$$

According to network disturbance concept as depicted in Figure 6, $\hat{d}$ can be also expressed as equation (18):

$$\hat{d} = u - uz^{-N} \quad (18)$$

where *u* is system input and *N* is the time delay.

In this way, the estimated disturbance $\hat{d}$ is used to compensate the time delay effect in the feedback signal.

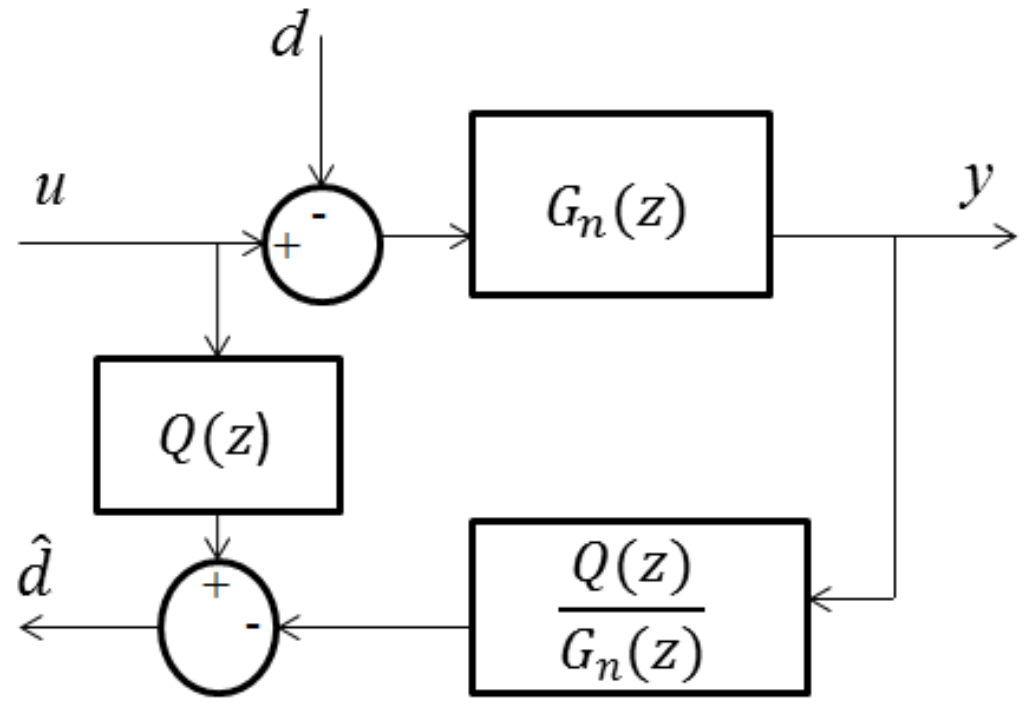

Figure 5. Classic disturbance observer in z domain

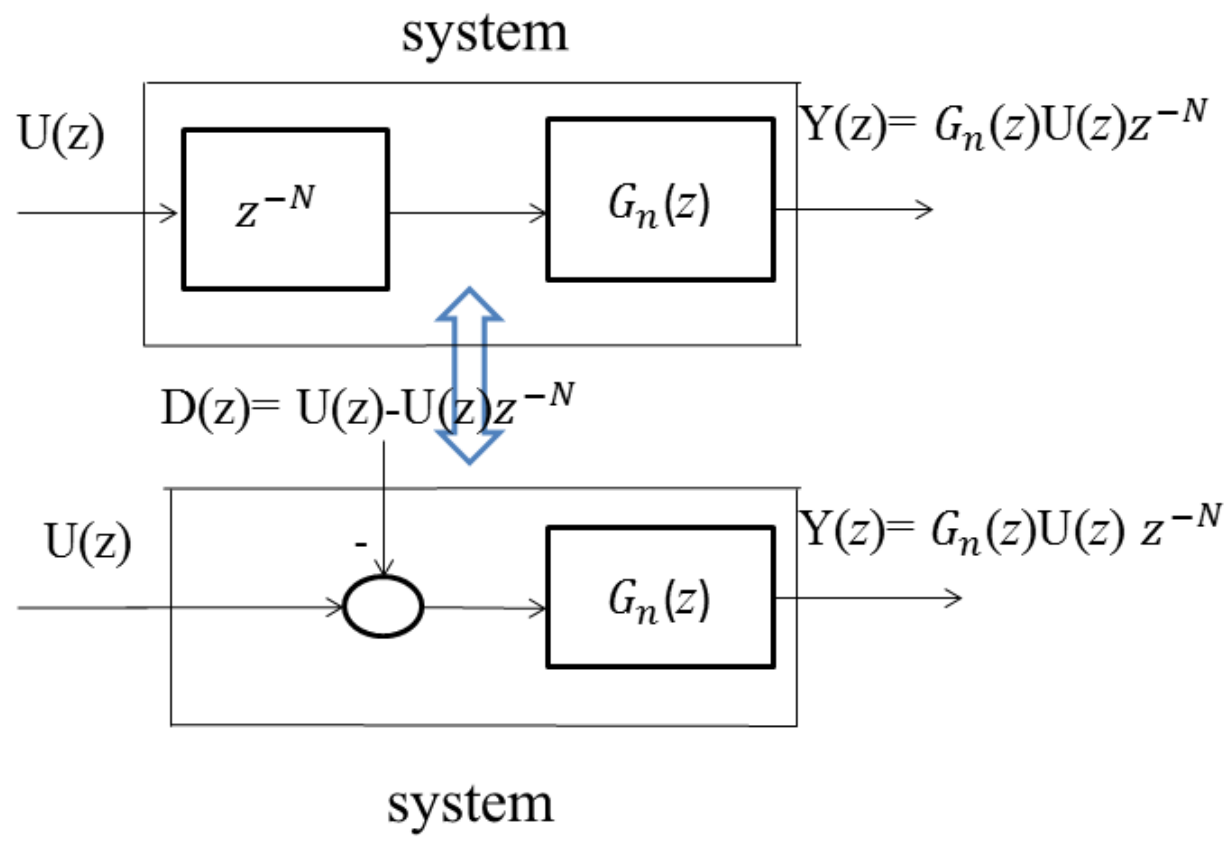

Figure 6. Conceptual block diagram of network disturbance

The block diagram of discrete-time CDOB compensated system is shown as Figure 7. Consider the time delay in the plant, $G(z) = G_n(z)z^{-N}$, $z^{-N}$ represents the time delay of discrete time vehicle model, where $G_n(z)$ is the nominal plant of $G(z)$. $C(z)$ is the digital robust PD controller which stabilizes the nominal plant $G_n(z)$. $Q(z)$ is the low pass filter. Based on the block diagram, the command regulation and disturbance rejection transfer functions can be derived in $z$ domain as equation (19) (20), respectively.

$$\frac{Y(z)}{R(z)} = \frac{CGz^{-N}}{1+CG_nQ+CGz^{-N}(1-Q)} \tag{19}$$

$$\frac{Y(z)}{D(z)} = \frac{1+CG_nQ}{1+CG_nQ+CGz^{-N}(1-Q)} \tag{20}$$

In CDOB, $Q$ filter is chosen as a low pass filter, with $Q \rightarrow 1$ and we can see that the time delay effect is eliminated from the closed-loop characteristic equation. In this way, the time delay is compensated by the CDOB.

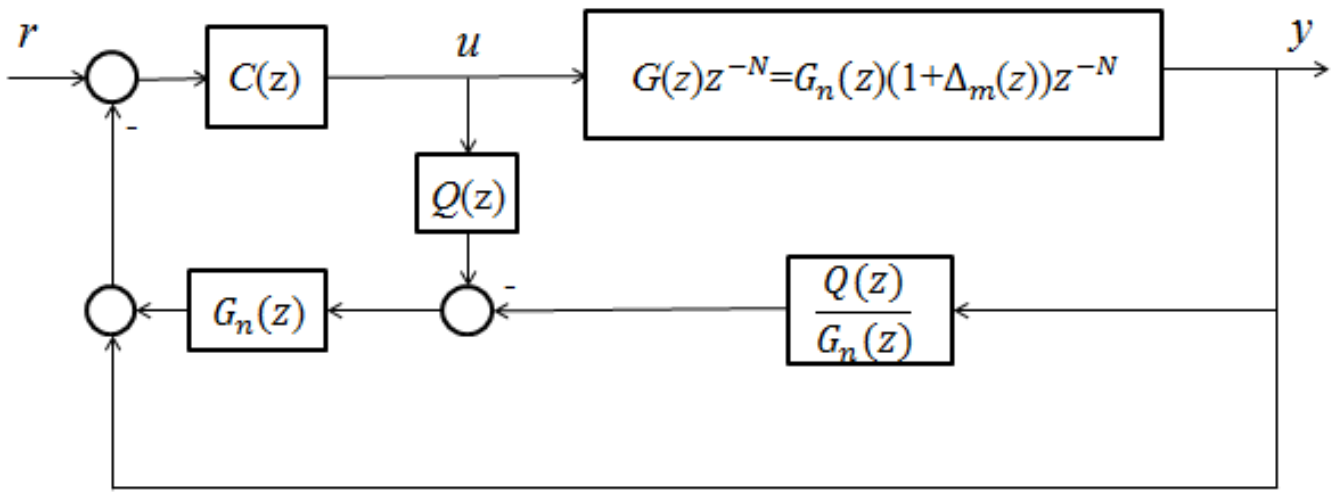

Figure 7. Digital CDOB compensated control system

## V. Digital Multi-Objective Robust PD Controller Design

Parameter space approach is developed from continuous time domain to discrete time domain for digital robust PD controller design. The details of parameter space approach based discrete time multi-objective robust PD controller design can be found in reference [21]. For digital multi-objective robust PD controller design, phase margin constraint and mixed sensitivity constraint are taken into account simultaneously. Phase margin is defined as $PM \in [20,80]$ deg and the parameters for mixed sensitivity constraint are: low frequency bound $l_s$ =0.5, the high frequency bound $h_s$ =4, and the approximate bandwidth is $\omega_s = 5\text{rad/sec}$ for sensitivity weight function $W_s$; low frequency gain $l_T$=0.2, the high frequency gain $h_T$=1.8, and the frequency of transition to significant model uncertainty $\omega_T = 120\text{rad/sec}$ for complementary sensitivity weight function $W_T$. Figure 8 illustrates the $k_d$-$k_p$ solution region and ($k_d$,$k_p$) design point is selected as (0.07, 0.2) as shown in red dot. It can be seen from Figure 9 that the corresponding frequency responses satisfy the phase margin constraint. Figure 10 shows the mixed sensitivity constraint is also satisfied with the chosen controller parameters as the magnitude plot is below 0dB $((|W_sS| + |W_TT|) = 1)$ line.

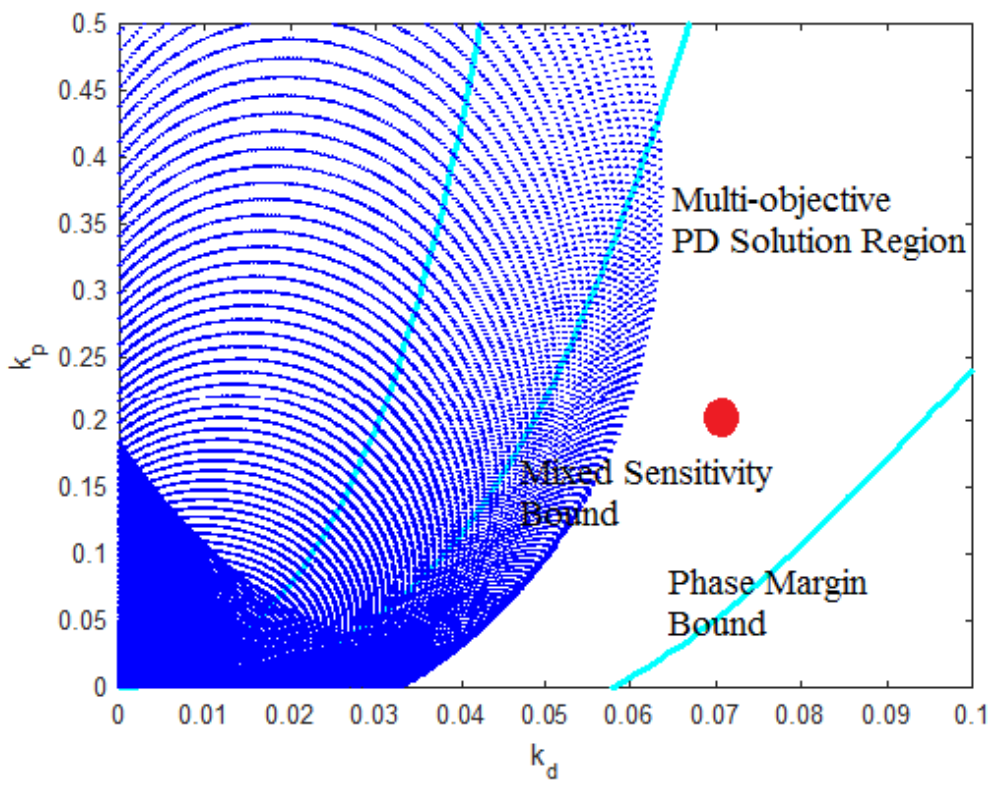

Figure 8. Multi-objective discrete time PD controller design

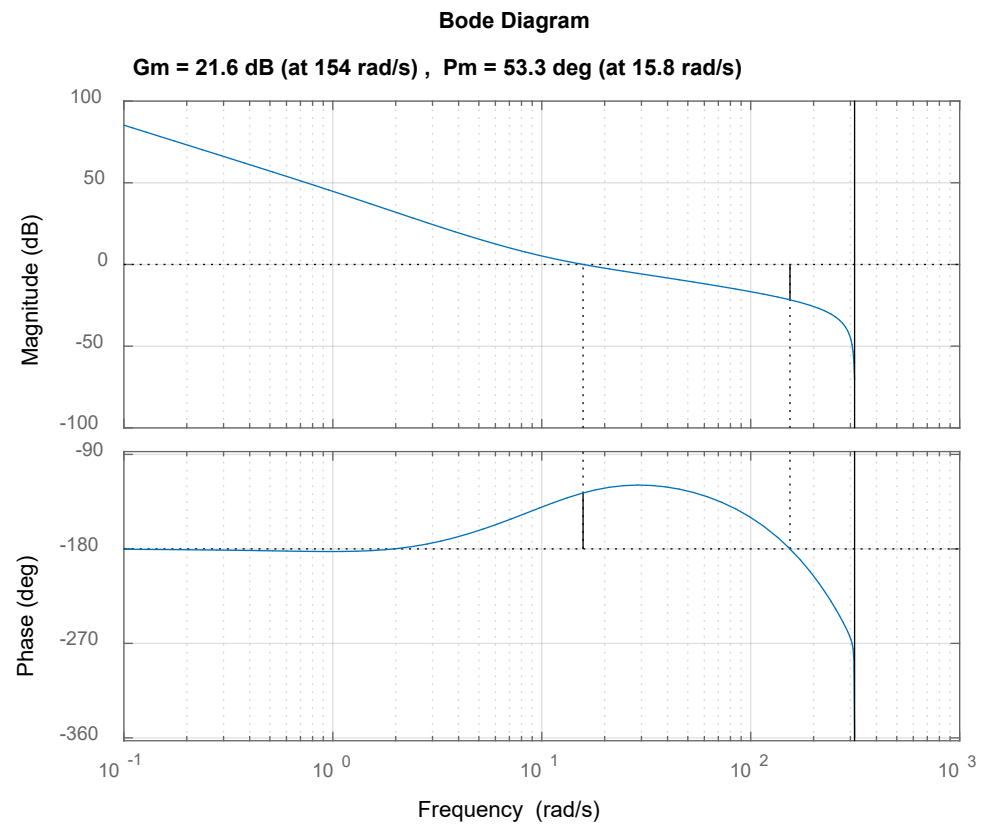

Figure 9. Phase margin constraint

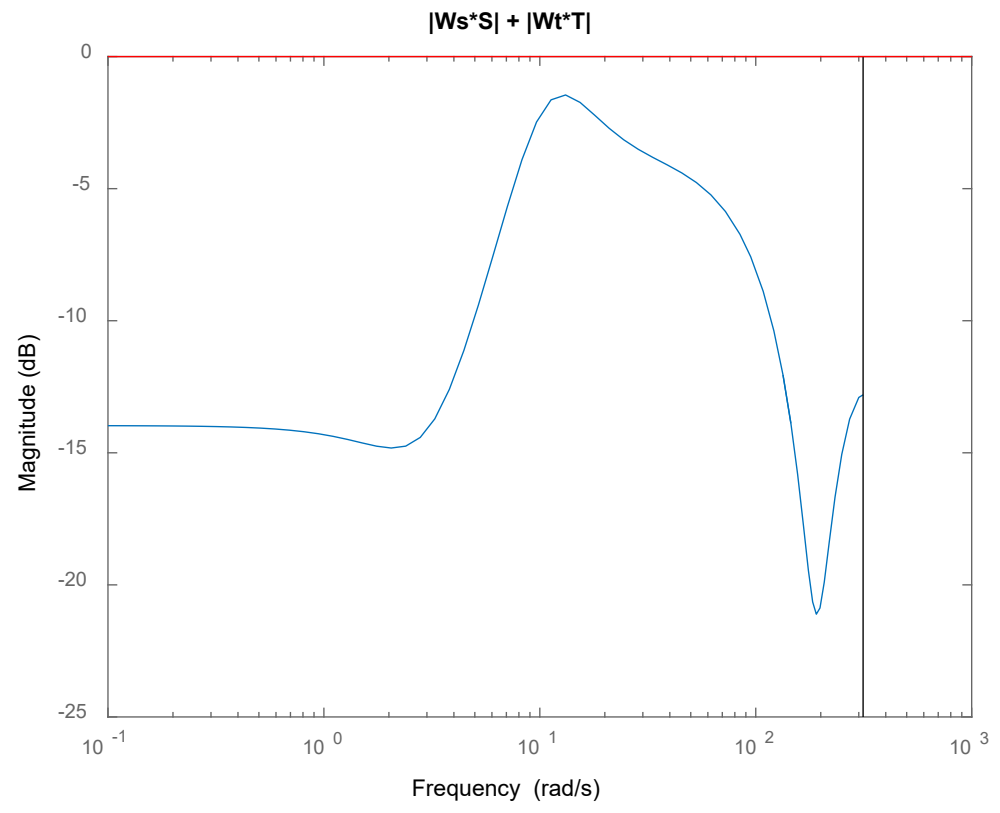

Figure 10. Robust performance plot

## VI. Discretization of DOB and CDOB Structure

Discrete time DOB structure and CDOB structure can be derived using zero-order-hold (ZOH) method. The nominal plant $G_n$ is the transfer function from front wheel steering angle $\delta_f$ to the lateral deviation $y$. According to Equation (5), the nominal plant $G_n$ in continuous time domain is calculated as:

$$G_n(s) = \frac{4713s^2 + 1.598\times10^5 s + 7.51\times10^5}{s^2(1.242 s^2 + 933.8s + 10610)} \quad (21)$$

For DOB structure, $Q_{DOB}$ is designed in our previous work [22] in continuous time domain as following equation:

$$Q_{DOB}(s) = \frac{1}{0.25s^2+s+1} \quad (22)$$

By using ZOH method, sampling time $T_s$ is given as 0.01 sec, $G_n(z)$ and $Q_{DOB}(z)$ are obtained as:

$$G_n(z) = \frac{0.04867 z^3 - 0.07432 z^2 + 0.02046 z + 0.005954}{z^4 - 2.892 z^3 + 2.784 z^2 - 0.8927 z + 0.0005429} \quad (23)$$

$$Q_{DOB}(z) = \frac{0.0001974z+0.0001974}{z^2 - 1.96z + 0.9608} \quad (24)$$

For CDOB structure, the cutoff frequency of $Q_{CDOB}(s)$ is determined as 50 rad/s [22] and $Q_{CDOB}(s)$ can be expressed as equation (25), discrete time $Q_{CDOB}(z)$ is derived as equation (26):

$$Q_{CDOB}(s) = \frac{1}{0.0004s^2+0.04s+1} \quad (25)$$

$$Q_{CDOB}(z) = \frac{0.0902 z + 0.06461}{z^2 - 1.213 z + 0.3679} \quad (26)$$

## VII. Simulation Studies

This section investigates autonomous vehicle path following performance of the proposed discrete time robust PD controlled system with DOB and CDOB compensation, respectively. An elliptical route is chosen as the desired path as shown in Figure 11 and the profile of 500 N crosswind force which acts as external disturbance is given in Figure 12. Vehicle initial position is at (0, 1) with 90° heading angle. The corresponding block diagrams for autonomous vehicle path following are given in Figure 4 and Figure 7, respectively. Matlab/Simulink is used as the simulation software. Sample time is 0.01 sec. Nominal plant $G(z)$ is given as equation (23) for both Figure 4 and 7. The designed digital robust PD feedback controller in Section V is applied as baseline controller for all simulations. In the DOB compensated closed loop control structure, low pass filter $Q(z)$ is determined as equation (24), and $Q(z)$ for robust PD controlled system with CDOB compensation is given as equation (26).

Figure 13-17 present the autonomous vehicle lateral deviation of discrete time robust PD feedback controller system with DOB compensation at four corners of the parametric uncertainty box and in the existence of crosswind disturbance, respectively. For comparison purpose, corresponding simulation results of PD only feedback control system are also shown in each figure. It can be seen that discrete time DOB compensated system has less path following errors compared with PD only feedback control system, which verifies that discrete time robust PD with DOB compensated system effectively deals with model regulation and disturbance rejection. Table II compares root-mean-square (RMS) errors of discrete time PD control with DOB compensation and discrete time PD control only, which also shows better path following performance of DOB structure.

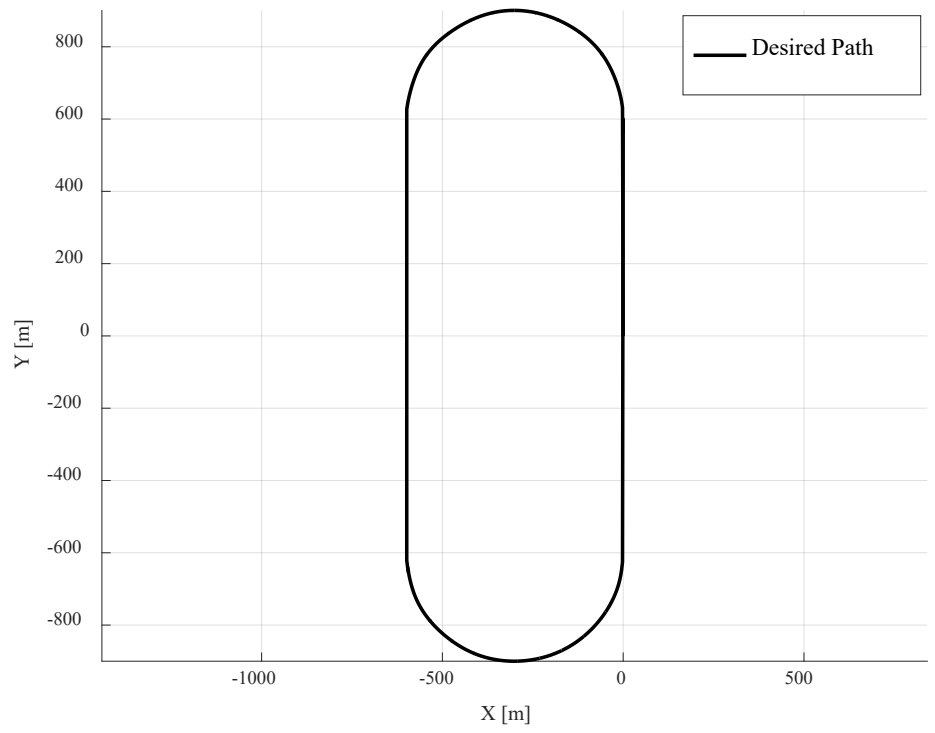

Figure 11. Desired path to be followed

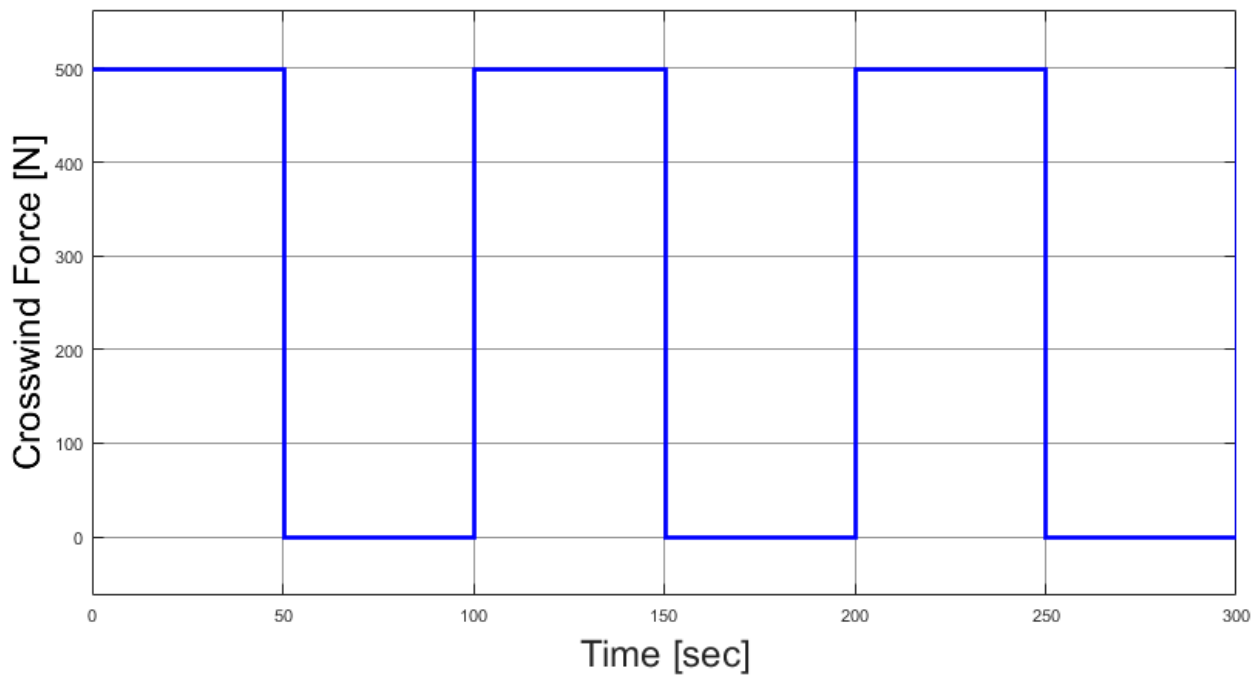

Figure 12. Crosswind force profile

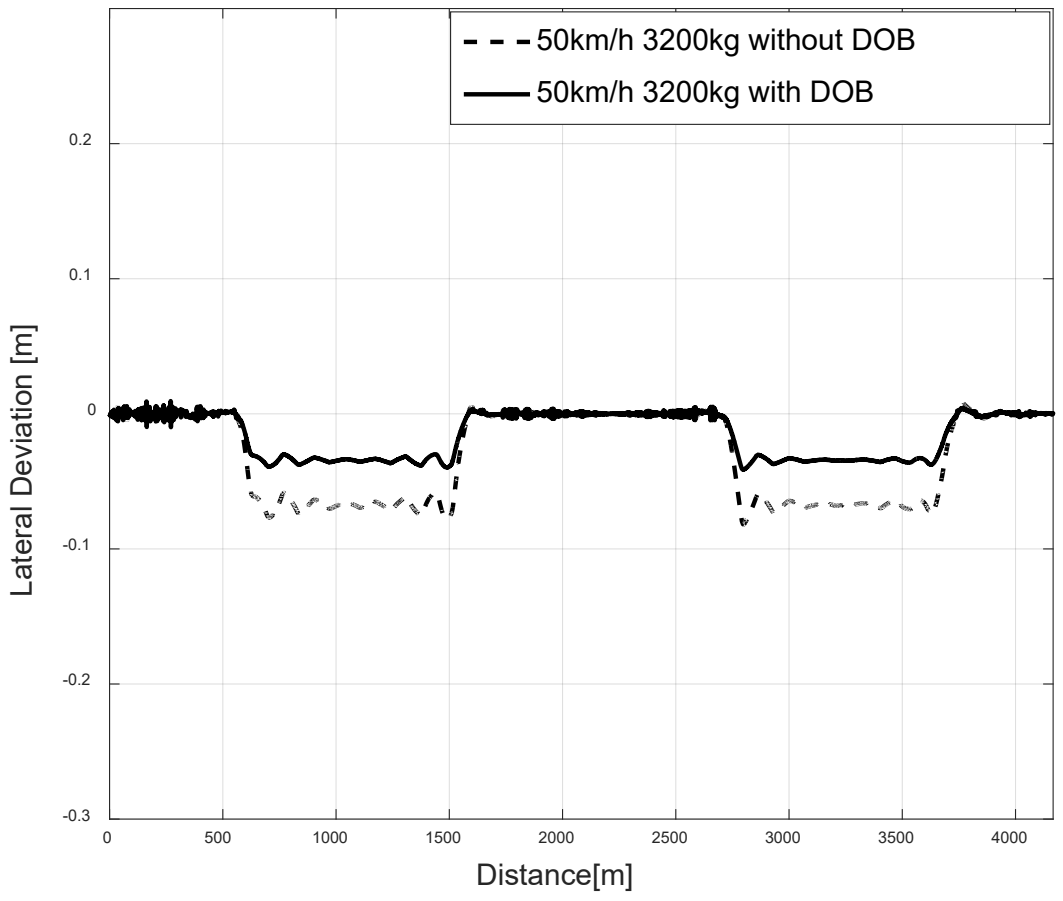

Figure 13. Lateral deviation with and without discrete DOB at corner *a*

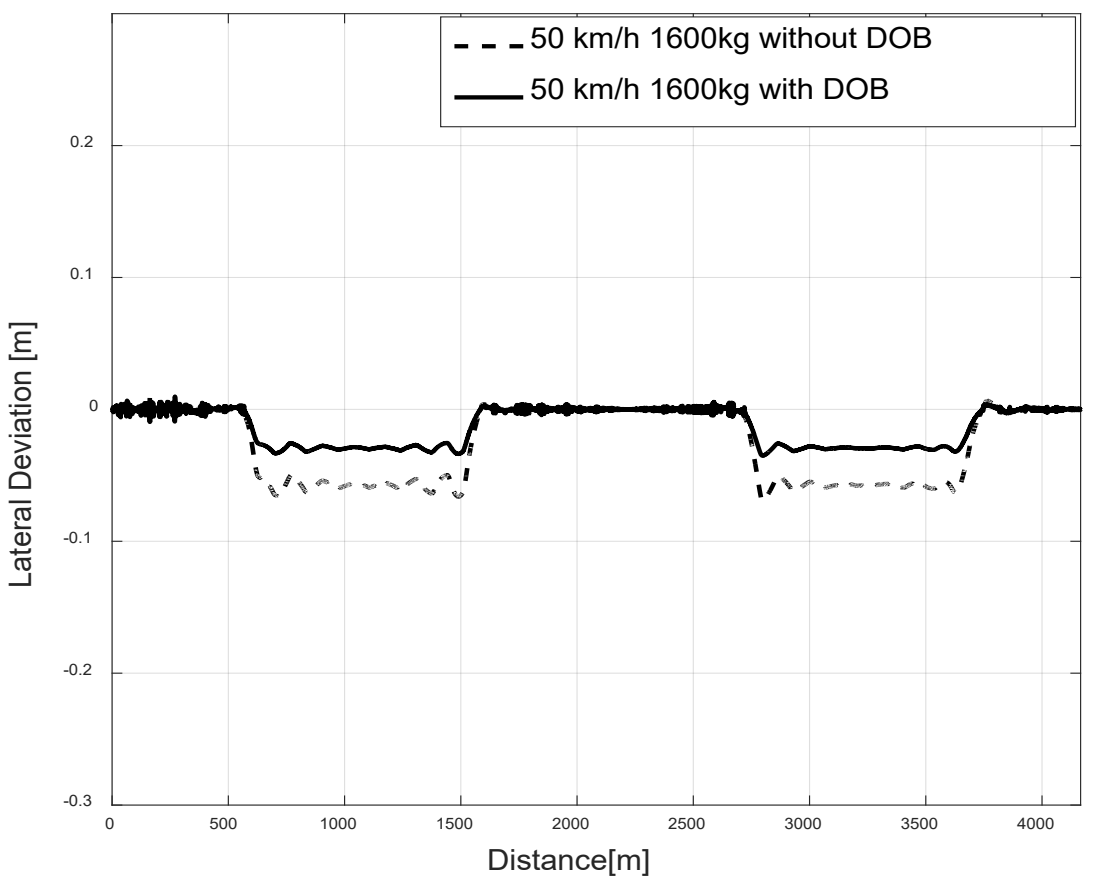

Figure 14. Lateral deviation with and without discrete DOB at corner *b*

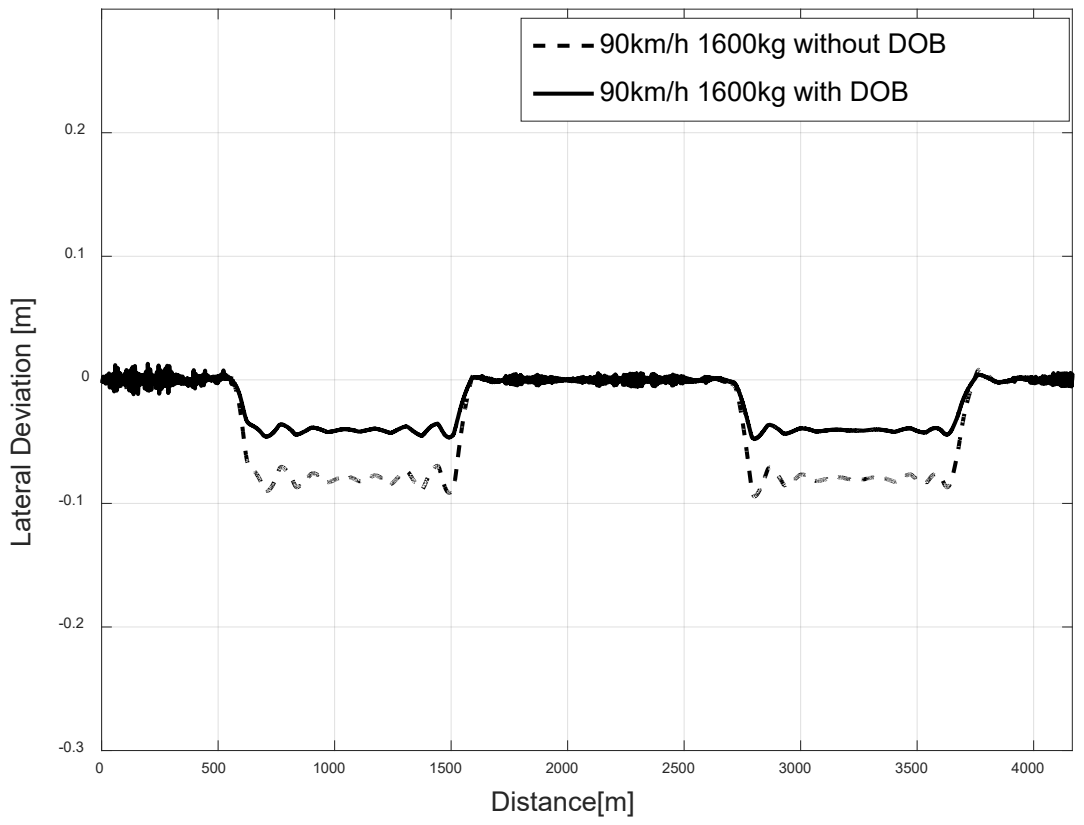

Figure 15. Lateral deviation with and without discrete DOB at corner *c*

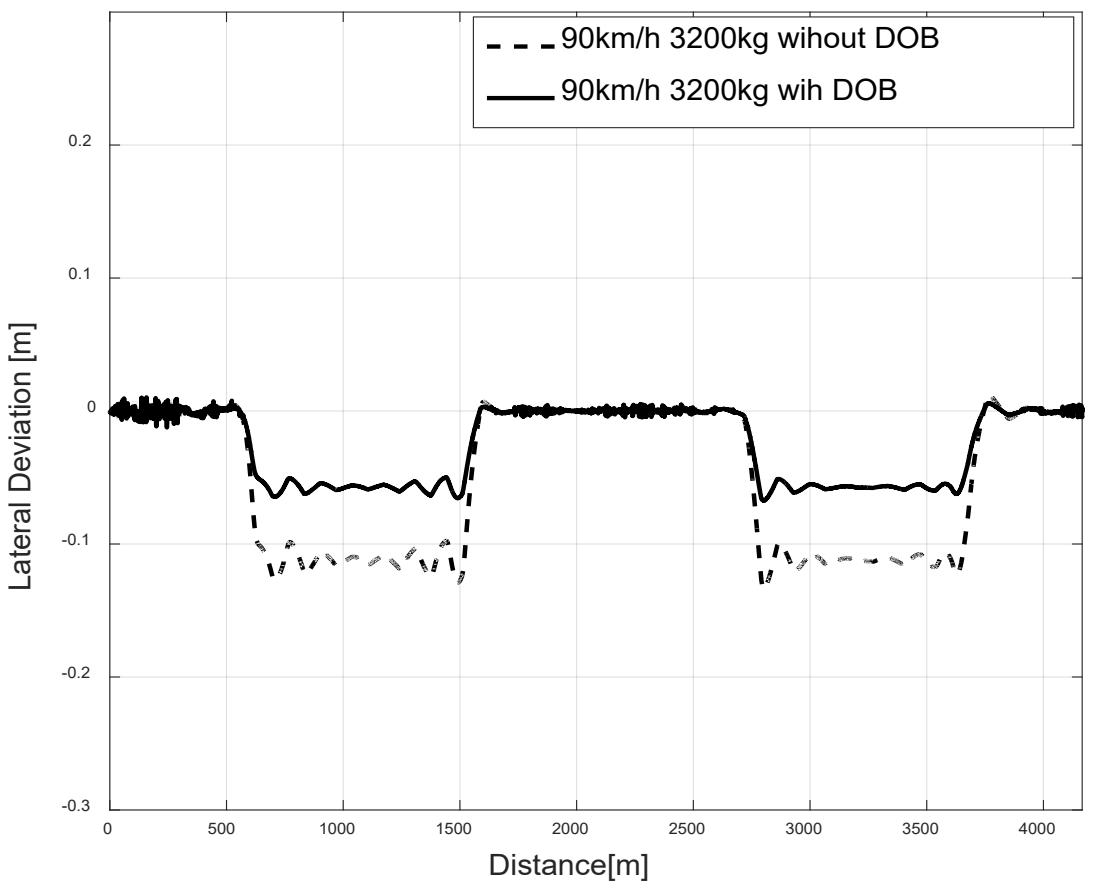

Figure 16. Lateral deviation with and without discrete DOB at corner *d*

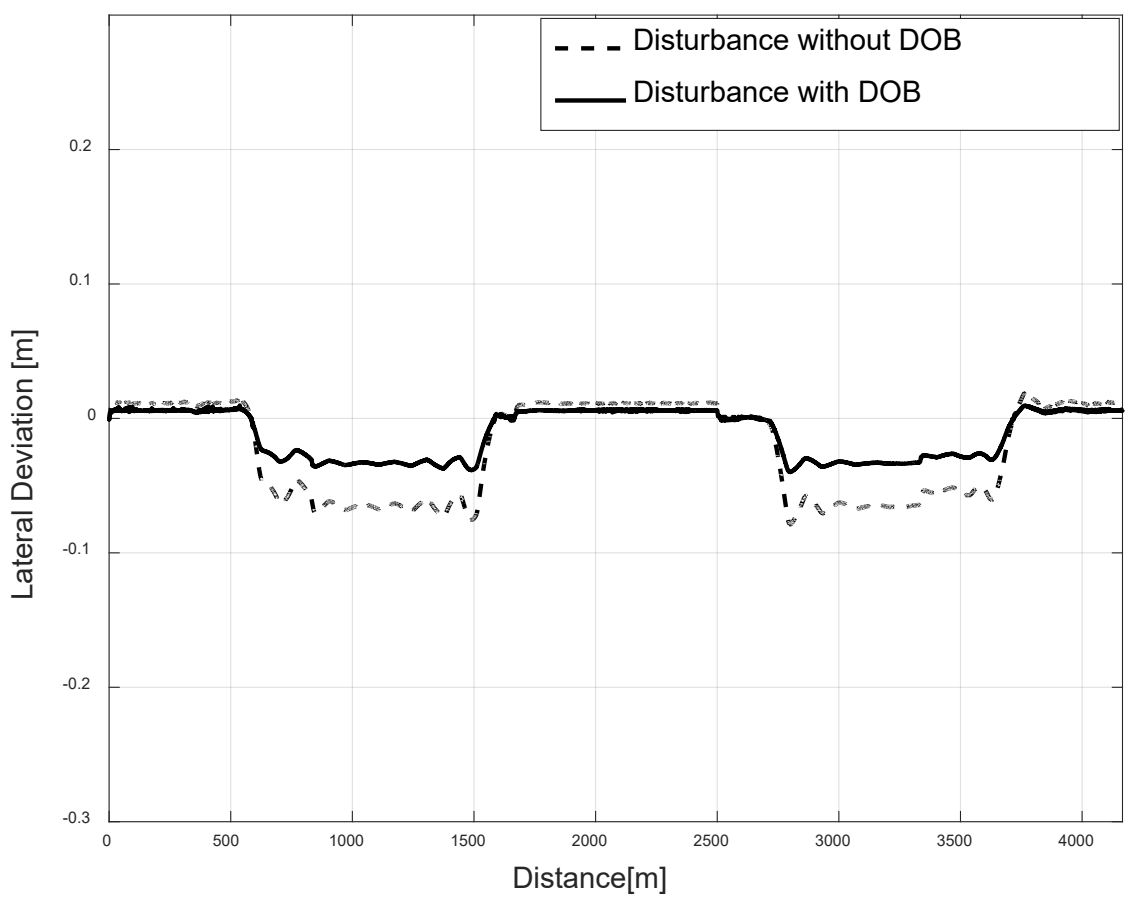

Figure 17. Lateral deviation with and without discrete DOB for crosswind input

Table II Comparison of RMS tracking errors between digital PD and digital PD with DOB

| Operating Condition / Control | 50km/h 1600kg | 50km/h 3200kg | 90km/h 1600kg | 90km/h 3200kg |
|---|---|---|---|---|
| PD | 0.0384m | 0.0451m | 0.0525m | 0.0739m |
| PD+DOB | 0.0196m | 0.023m | 0.0268m | 0.0377m |

Figure 18 compares the autonomous vehicle lateral deviation of discrete time robust PD feedback controlled system with and without communication disturbance observer compensation by taking time delay $N$=1 sec into account. It is shown in Figure 18 that in the existence of time delay, PD feedback control system is not stable and oscillates, while CDOB compensation stabilizes the system and its lateral deviation is close to the one from nominal plant. Figure 19 presents the lateral deviation of CDOB compensated system for different time delay values. It is seen that path following errors do not increase with the increase of delay time. It can be concluded that discrete time CDOB compensates time delay and improves autonomous vehicle path following performance.

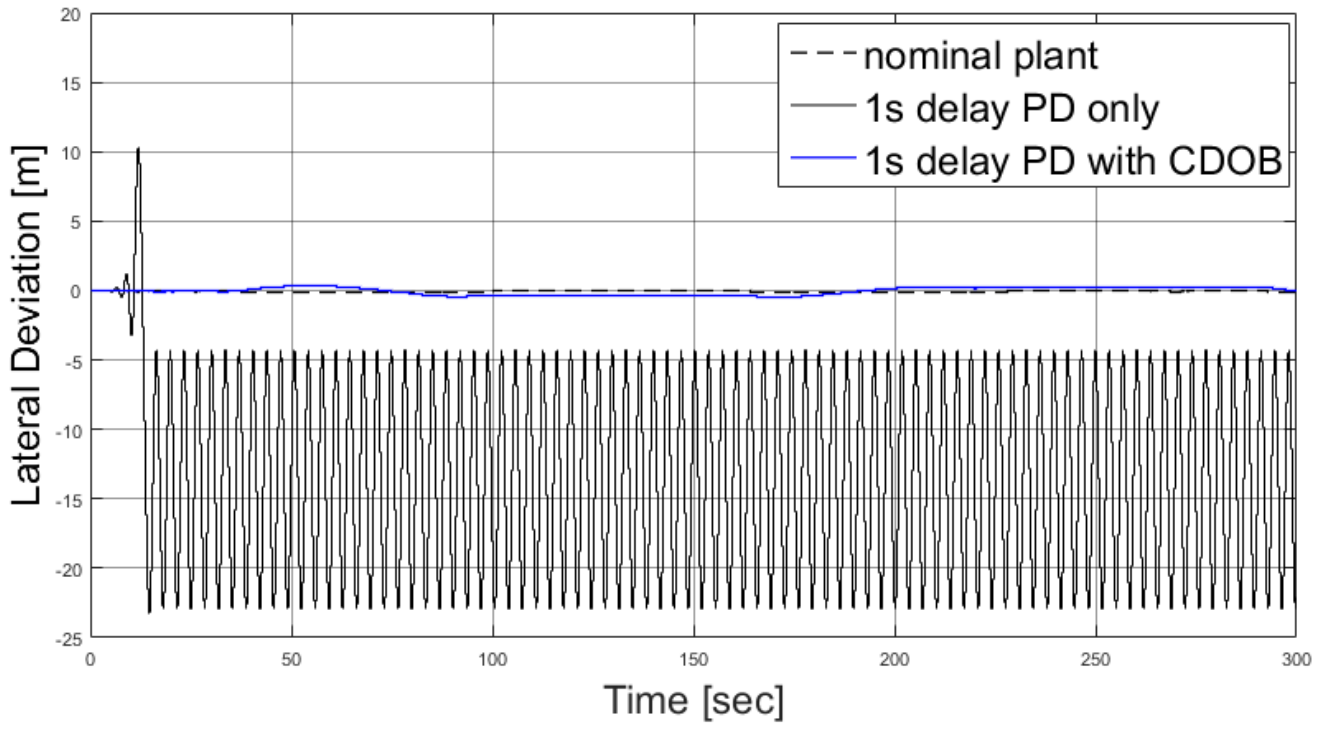

Figure 18. Lateral deviation with and without CDOB for time delay

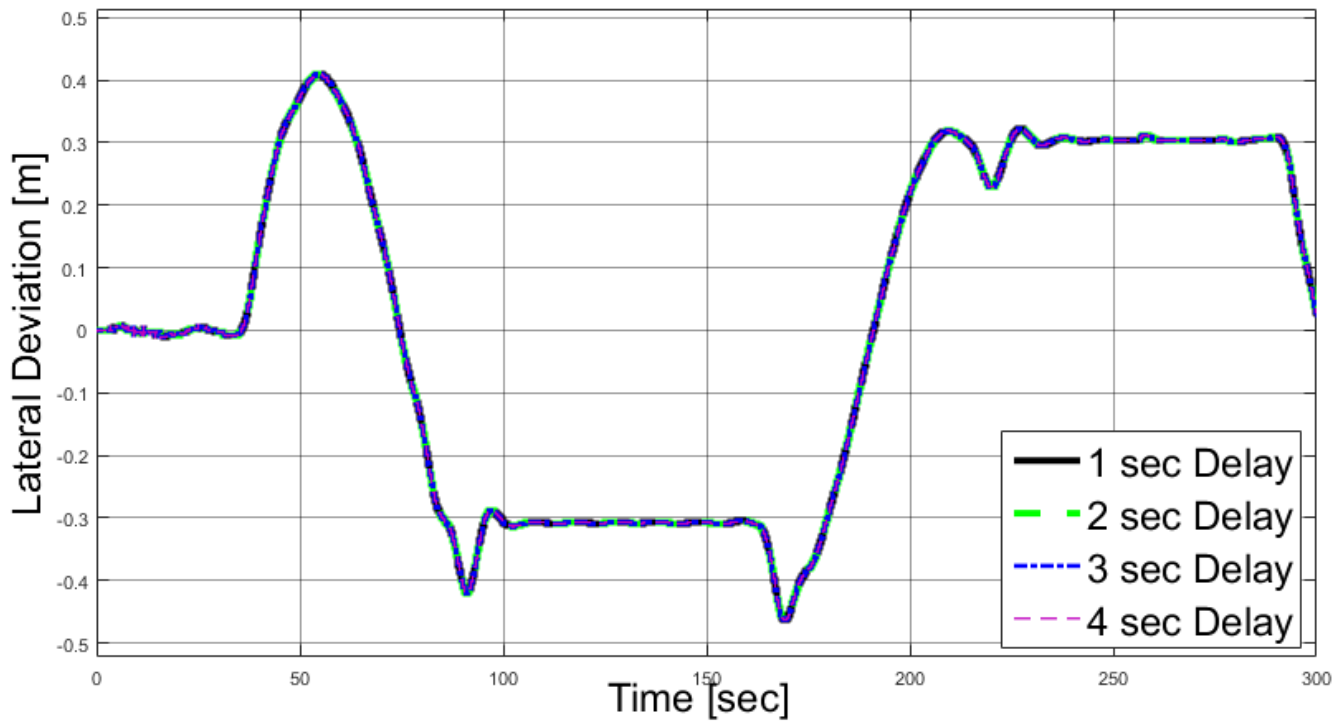

Figure 19. Lateral Deviations of CDOB compensation for different time delay

# VIII. Conclusion

This paper applies robust PD feedback control with DOB/CDOB compensated system in discrete time domain and simulations are carried out to verify the effectiveness of the proposed structure. Digital robust PD controller is designed based on parameter space approach. Discrete-time DOB and CDOB structures are obtained by discretizing continuous time DOB and CDOB structures using zero order hold method. Autonomous vehicle path following driving at high speed is performed to validate the proposed discrete time DOB and CDOB structure. Simulation results show that the proposed discrete time DOB structure realizes model regulation and disturbance rejection, and discrete time CDOB effectively deals with time delay compensation. These prove the successful implementation of robust PD with DOB and CDOB compensated system in discrete domain. In the future work, hardware-in-the-loop simulations and experiments will be performed to further test the designed discrete-time DOB and CDOB systems. Future work can also focus on and treat integrating or combining some of the approaches in this paper with other control, AV, CV, ADAS, automotive control and other topics like those in references [23-84] and others in the future.

## Acknowledgments

This paper is based upon work supported by the National Science Foundation under Grant No.:1640308 for the NIST GCTC Smart City EAGER project UNIFY titled: Unified and Scalable Architecture for Low Speed Automated Shuttle Deployment in a Smart City, by the U.S. Department of Transportation Mobility 21: National University Transportation Center for Improving Mobility (CMU) sub-project titled: SmartShuttle: Model Based Design and Evaluation of Automated On-Demand Shuttles for Solving the First-Mile and Last-Mile Problem in a Smart City.